\documentclass[sigconf,natbib=false]{acmart}
\acmConference[CAIN 2024]{3rd International Conference on AI Engineering — Software Engineering for AI}{April 2024}{Lisbon, Portugal}

\usepackage{algorithmic}
\usepackage{graphicx}
\usepackage{textcomp}
\usepackage{hyperref}
\usepackage{paralist}
\usepackage{multirow}
\usepackage{xcolor}

\AtBeginDocument{%
  }

\setcopyright{acmcopyright}
\copyrightyear{2018}
\acmYear{2018}
\acmDOI{XXXXXXX.XXXXXXX}

\RequirePackage[
  datamodel=acmdatamodel,
  style=acmnumeric,
  ]{biblatex}

\copyrightyear{2024}
\acmYear{2024}
\setcopyright{rightsretained}
\acmConference[CAIN 2024]{Conference on AI Engineering Software Engineering for AI}{April 14--15, 2024}{Lisbon, Portugal}
\acmBooktitle{Conference on AI Engineering Software Engineering for AI (CAIN 2024), April 14--15, 2024, Lisbon, Portugal}\acmDOI{10.1145/3644815.3644961}
\acmISBN{979-8-4007-0591-5/24/04}

\begin{document}

\title{Is Your Anomaly Detector Ready for Change? Adapting AIOps Solutions to the Real World}

\author{Lorena Poenaru-Olaru}
\email{L.Poenaru-Olaru@tudelft.nl}
\affiliation{%
  \institution{Delft University of Technology}
  \country{Netherlands}
}

\author{Natalia Karpova}
\email{karp.n.b@gmail.com}
\affiliation{%
  \institution{Delft University of Technology}
  \country{Netherlands}
}

\author{Luis Cruz}
\email{L.Cruz@tudelft.nl}
\affiliation{%
  \institution{Delft University of Technology}
  \country{Netherlands}
}

\author{Jan S. Rellermeyer}
\email{rellermeyer@vss.uni-hannover.de}
\affiliation{%
  \institution{Leibniz University Hannover}
  \country{Germany}
}

\author{Arie van Deursen}
\email{arie.vandeursen@tudelft.nl}
\affiliation{%
  \institution{Delft University of Technology}
  \country{Netherlands}
}

\renewcommand{\shortauthors}{Poenaru-Olaru et al.}

\begin{abstract}
Anomaly detection techniques are essential in automating the monitoring of IT systems and operations. These techniques imply that machine learning algorithms are trained on operational data corresponding to a specific period of time and that they are continuously evaluated on newly emerging data. Operational data is constantly changing over time, which affects the performance of deployed anomaly detection models. Therefore, continuous model maintenance is required to preserve the performance of anomaly detectors over time. In this work, we analyze two different anomaly detection model maintenance techniques in terms of the model update frequency, namely blind model retraining and informed model retraining. We further investigate the effects of updating the model by retraining it on all the available data (full-history approach) and only the newest data (sliding window approach). Moreover, we investigate whether a data change monitoring tool is capable of determining when the anomaly detection model needs to be updated through retraining.

\end{abstract}



\maketitle

\section{Introduction}
The field of AIOps refers to applying artificial intelligence (AI) techniques on large-scale operational data to solve challenges derived from operational workflows. AIOps aims to increase the productivity of IT Ops, DevOps, and software reliability engineering teams by predicting the behavior of large-scale software systems and improving the software architectural decision-making processes~\cite{aiopsrwchallenges}. The term AIOps solutions is used to describe machine learning (ML) systems that learn from operational data. In the past years, AIOps solutions have witnessed a fast adoption within different industries. The most popular AIOps solutions are failure prediction and anomaly detection~\cite{aiopsovertime}.

Given that any AIOps solution is an ML system, the quality of its predictions is strongly dependent on the data it was trained on and the data it is evaluated on after being deployed into production. If the training data is significantly different than the evaluation data, there is a high chance that the performance of the ML system will be affected~\cite{lorena2022concept}. Previous work on failure prediction AIOps solutions has observed the evolving character of operational data, which implies that this data is continuously changing over time~\cite{datasplittingdecisions},~\cite{operationaldatachanging1},~\cite{nodefailureretrain},~\cite{nodefailureretrain2}. The continuous data changes, also known as concept drift, influence the performance of AIOps solutions during their lifecycle. They become less accurate over time, which tremendously affects their reliability among practitioners. Although the concept drift cannot be prevented since it is caused by external hidden factors, AIOps practitioners need to constantly ensure that their AIOps solutions are up to date.




One solution that ML practitioners adopt to handle the evolving character of data is retraining/updating ML models over time~\cite{nadiaicse}. Periodical model retraining has also been studied for failure detection AIOps solutions~\cite{towardsconsistentinterpretationofAIOps},~\cite{datasplittingdecisions} and has proved that continuous model updates achieve better performance over time compared to non-updated models. However, the effects of continuously updated models have only been studied for failure prediction models~\cite{datasplittingdecisions},~\cite{towardsconsistentinterpretationofAIOps}.

In ~\cite{lorenamonitoringmaintaining} the authors proposed a solution to handle concept drift in failure prediction AIOps solutions, which implies that the model is updated only when the concept drift is detected, instead of periodically. Therefore, their solution contains a monitoring part responsible for identifying concept drift in the evaluation data using a concept drift detector. However, this framework was just proposed, but it was never assessed on any AIOps solution. 

When it comes to periodic model retraining, this technique has been studied for classification problems such as failure predictions AIOps solutions. Lyu et al.~\cite{towardsconsistentinterpretationofAIOps} suggested that it also needs to be studied on other AIOps solutions, such as anomaly detection on time series data. By doing so, AIOps practitioners could have a better understanding of whether they can mitigate the effects of concept drift by constantly updating anomaly detectors. Regarding concept drift monitoring-based model retraining, this technique was not previously applied to any AIOps solution. Furthermore, previous work suggests that organizations do not have monitoring infrastructure to detect drift in production~\cite{nadiaicse},~\cite{nadiacain} and they only perform periodic model retraining based on human decisions. Examining the impact of drift detection monitoring tools is the first step toward automating the maintenance of machine learning in production. Therefore, it could help in understanding whether concept drift detectors could be quality and reliability indicators for deployed ML models.

To mitigate the effects of concept drift on anomaly detection AIOps solutions, in this paper, we study different model adaptation techniques. Our contributions can be summarized as follows:
\begin{compactenum}

    \item We examine the effect of periodically updating models on the performance of anomaly detection AIOps solutions.
    \item We assess and report the limitations of the state-of-the-art anomaly detection models when being evaluated on different data sizes. 
    \item We investigate the effects of retraining anomaly detection AIOps models based on the output of a concept drift detector.
    \item A publicly available replication package is provided including the implementation of a concept drift detector for time series.
\end{compactenum}


\section{Background and Related Work}
\subsection{Background AIOps Solutions \& Anomaly Detection}


AIOps solutions aim to identify issues in large software systems and then help with mitigating them or providing recommendations to engineers. When it comes to detecting possible issues in the system, a variety of different AIOps solutions were proposed for failure prediction tasks, namely predicting job failure \cite{jobfailureprediction}, node failure \cite{nodefailureprediction}, disk failure \cite{diskfailurepredicrion}, incident \cite{incidentprediction} or outage \cite{outageprediction}. Besides failures, plenty of attention has been paid to identifying abnormal system behavior, such as performance anomalies~\cite{performanceanomalies}, \cite{performanceanomalies2}, anomalies in system logs~\cite{loganomalydetect1},~\cite{loganomalydetect2}, or internet traffic anomalies \cite{anomalyDetectionMicrosoft}.


Although plenty of anomaly detection techniques were proposed in the literature, regarding AIOps solutions for anomaly detection, previous work has focused chiefly on \textit{unsupervised} and \textit{semi-supervised} models. This is a technique that practitioners use to handle the label availability challenge and high cost of obtaining true labels~\cite{anomalyDetectionMicrosoft}. The best-performing and most popular techniques to detect anomalies in univariate AIOps data~\cite{ajointmodelforit},~\cite{anomalyDetectionMicrosoft},~\cite{anomalydetectionAlibaba} belong to the \textit{signal reconstruction models} group~\cite{anomalysurvey2}. 

Anomalies are detected using methods that encode the time series into a latent space, such as Fast \textit{Fourier Transform (FFT)}~\cite{fft}, \textit{Spectral Residuals (SR)}~\cite{anomalyDetectionMicrosoft} or \textit{Prediction
Confidence Interval (PCI)}~\cite{pci}. These techniques usually have low computational costs, but they lose information during the encoding process~\cite{anomalysurvey2}. To preserve more information, a type of artificial neural network called Auto-Encoder (AE) is employed to transform the time series data into a latent space. Thereby, plenty of anomaly detection techniques based on Auto-Encoders were derived, namely \textit{Long Short-Term Memory
Autoencoder (LSTM-AE)}~\cite{lstmae} or \textit{DONUT}~\cite{vae}. Furthermore, Microsoft presents a more complex anomaly detector, \textit{Spectral Residuals Convolutional
Neural Networks (SR-CNN)}~\cite{anomalyDetectionMicrosoft} that learns from multiple time series, generates synthetic anomalous samples, and trains a Convolutional Neural Network (CNN) to distinguish between anomalous and non-anomalous samples.

The anomaly detection models were previously evaluated using a delay metric. The reason for this is that in real-world applications, anomalies can occur either as single points or as segments of anomalies (group of continuous anomalies) \cite{groupvspointanomaly}. According to \cite{anomalyDetectionMicrosoft}, \cite{unsupervisedADviaVAE}, \cite{ajointmodelforit}, in AIOps solutions detecting any anomaly point in a segment of anomalies with a relatively small delay is considered as successful as detecting all anomaly points belonging to the same segment. Therefore, a delay-based evaluation strategy was presented in previous anomaly detection studies \cite{anomalyDetectionMicrosoft}, \cite{unsupervisedADviaVAE}, \cite{ajointmodelforit}.

\subsection{Concept Drift in Operational Data and Model Adaptation Techiques}


Previous work observed the evolving character of operational data, which is responsible for changes over time and, therefore, for concept drift~\cite{datasplittingdecisions},~\cite{aiopsrwchallenges},~\cite{nodefailureretrain},~\cite{nodefailureretrain2}. 
The presence of concept drift results in the degradation in performance of the failure prediction models. Therefore, AIOps solutions need to be constantly maintained over time by continuous retraining~\cite{datasplittingdecisions},~\cite{towardsconsistentinterpretationofAIOps}.

Gama et al.~\cite{conceptdriftadaptationsurvey} propose two machine learning retraining techniques from the perspective of the retraining frequency, namely \textit{blind retraining} and \textit{informed retraining}. Blind retraining is the equivalent of periodic retraining, where the model is retrained after a predefined period. This technique was previously used by previous work on failure prediction models~\cite{nodefailureretrain},~\cite{nodefailureretrain2},~\cite{datasplittingdecisions},~\cite{towardsconsistentinterpretationofAIOps},~\cite{modelretraining3} and proved to be beneficial for preserving the model's performance over time. Informed retraining implies the existence of a data monitoring tool called a \textit{concept drift detector} which indicates when the model is outdated due to changes in data. However, there is currently no study on the effects of blind and informed retraining on anomaly detection models.

Lyu et al.~\cite{towardsconsistentinterpretationofAIOps} propose two retraining techniques from the perspective of retraining data, namely the \textit{full-history approach} and the \textit{sliding window approach}. The full-history approach constantly enriches the training dataset with the newest data and retrains the model, while the sliding window approach retrains the model only on the most recent data, discarding old samples. These methods have been studied in failure prediction models, but there is currently no work on the effects of the full-history approach and sliding window on anomaly detection models.

\subsection{Concept Drift Detection}

Concept drift is monitored using some algorithms that can capture the moment when data shift occurs called \textit{concept drift detectors}~\cite{lorena2022concept}. Although there are plenty of drift detectors available for multivariate data used in classification purposes when it comes to time series, the number of available drift detection techniques is significantly lower~\cite{BAYRAM2022108632}. According to Bayram et al.~\cite{BAYRAM2022108632}, the reason for this is the lack of available open-source relevant time series datasets to study drift detection. Despite the potential of concept drift detectors to monitor data against concept drift, they have not been yet applied to the AIOps domain. Furthermore, there is no study on concept drift monitoring techniques for anomaly detection on operational data.

Out of the existing concept drift detectors for time series we can mention \textit{Feature Extraction Drift Detection (FEDD)}~\cite{fedd} and \textit{Entropy-Based Time Domain Feature Extraction (ETFE)}~\cite{etfe}. Both these detectors identify drift by extracting features from a given time series window and observing their similarity with the features extracted from the reference window. FEDD is extracting six linear (autocorrelation, partial autocorrelation, variance, skewness coefficient, kurtosis coefficient, and turning point rate) and two non-linear features (bicorrelation and mutual information) and computing the dissimilarity between features extracted in the current window and the reference window. The drift is detected through a change detector which analyses the exponentially weighted moving average (EWMA) of the computed dissimilarities. ETFE is extracting solely entropy-related features (approximate, fuzzy, sample, permutation, increment, and weighted permutation entropy) from the decomposed time series. Concept drift is detected from the features extracted from the reference and the current time series using the GLR statistical test.

\section{Motivational Example}

\begin{figure}
    \centering
    \includegraphics[width=0.5\textwidth]{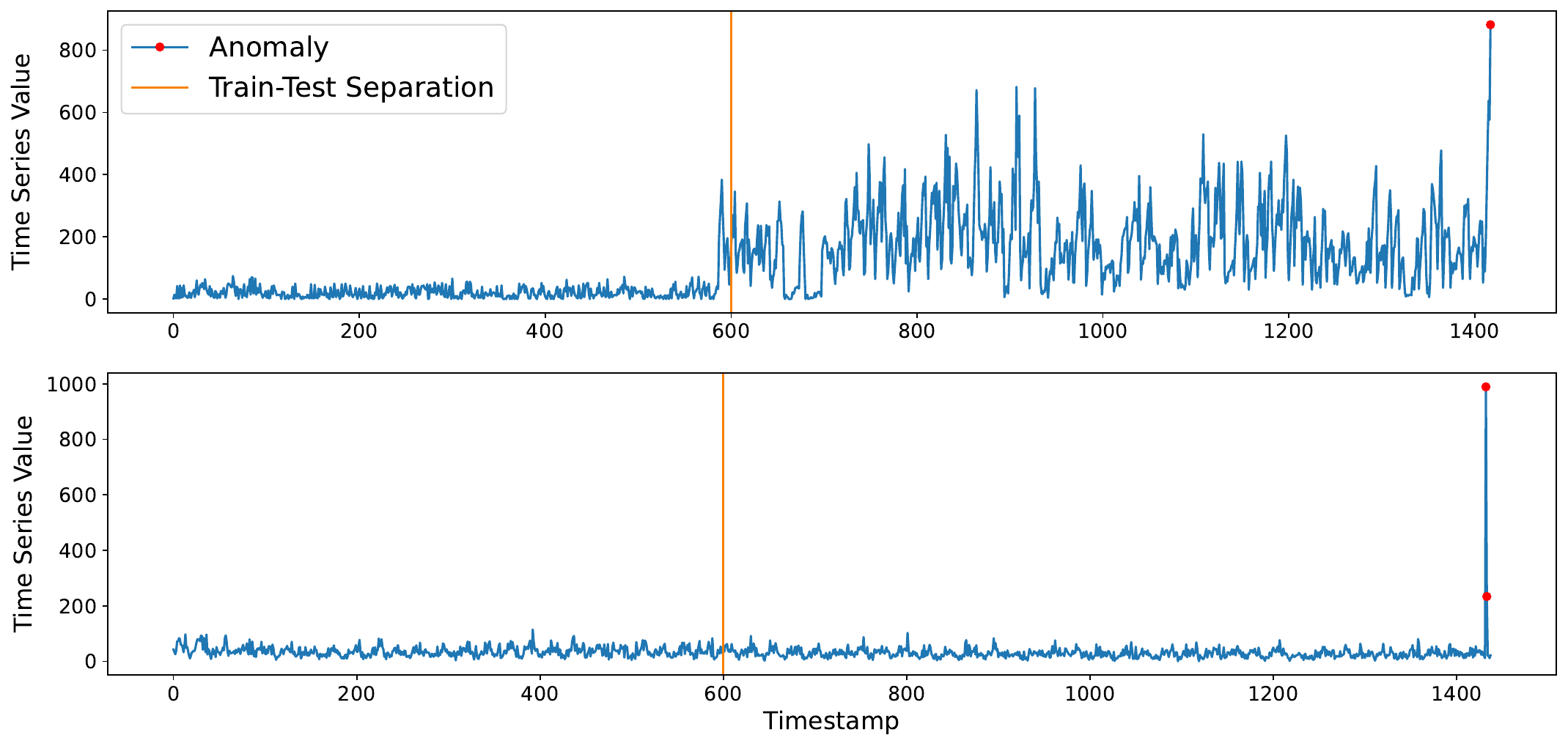}
    \caption{The time series to the left of the vertical line represents the training data, while the one to the right represents the evaluation/testing data.}
    \label{figure:motivational_example}
    \vspace{-1em}
\end{figure}

In Figure \ref{figure:motivational_example}, we depict an example of two different time series related to internet traffic from the Yahoo dataset, a popular benchmark for AIOps data related to internet traffic~\cite{anomalydetectionAlibaba},~\cite{anomalyDetectionMicrosoft},~\cite{ajointmodelforit}. In the upper plot, the behavior of operational data is significantly changing after a certain timestamp, while in the lower plot, it remains relatively constant. Thus, different time series have different behaviors over time. 

We assume that we train two anomaly detection models on the data corresponding to the left of the vertical line and we keep the rest for testing. In the upper time series, we notice that the testing data is significantly different than the training data, which means that the model requires updating over time. Otherwise, the model would signal plenty of false alarms, which need to be verified by the operation and service engineers. Thus, the advantage of saving data monitoring time using anomaly detection AIOps solutions is diminished by the amount of time engineers need to spend
searching for the causes of erroneously identified anomalies. Furthermore, the engineers would doubt the reliability of the model and they will be more reluctant to employ it. However, we notice that the data of the lower time series is not changing over time, thus the model should be able to capture anomalies.

If we would maintain our models using \textit{blind retraining}, the model corresponding to the lower time series would be retrained although their performance might not be drastically improved since there were no changes in the data. Thus, we might encounter unnecessary retraining costs, which could be avoided by employing \textit{informed retraining} where we initially verify whether data has changed and then update the model. When it comes to the model corresponding to the upper time series, both blind retraining and informed retraining might be beneficial. 

Using a full-history approach in the case of the upper time series might result in poor anomaly detection performance for the upper time series since the past data no longer resembles current data. However, in the case of the lower time series, both techniques sliding window and full-history approaches might lead to similar results. 

Given that time series are different, they might require different maintenance techniques. Therefore, in our study, we address the effects of updating different models over time on their performance and the implications of each maintenance technique in terms of how accurately they detect anomalies and how many false alarms they raise.

\section{Research Questions}

We begin by assessing the performance of state-of-the-art anomaly detection models on operational data. Moreover, we aim to understand whether the size of the testing set affects the performance of these models. We further aim to investigate the lifecycle of these models in the situation in which the models are maintained over time compared to when they are never updated. We analyze these aspects by evaluating the model retraining effects from the perspective of the retraining data and retraining frequency.
\begin{compactenum}
    \item What is the performance of state-of-the-art anomaly detection models on operational data?
    \begin{compactenum}
    \item How robust are the state-of-the-art models to the testing set size?
    \end{compactenum}
    \item What is the impact of the two model retraining techniques from the perspective of the retraining data (full-history vs sliding window approach)?
    \item What is the impact of the two model retraining techniques from the perspective of the retraining frequency (blind vs informed retraining)?
\end{compactenum}

\section{Evaluation Methodology}

In this section, we present the evaluation methodology that we employ to answer our research questions. We begin by describing the employed datasets and anomaly detection models. We continue by presenting different retraining (maintenance) techniques and the concept drift detector monitoring tool that signals changes in the data over time. Lastly, we explain the used evaluation metrics together with the delay tolerance.

\subsection{Datasets}

Although there are various benchmarks for anomaly detection, very few benchmarks for AIOps anomaly detection exist since research on AIOps is mostly performed using proprietary production data that is not publicly released~\cite{aiopsarxiv}. Therefore, we selected two popular operational datasets containing univariate time series with different data collection granularity and lengths as presented in Table \ref{table_datasets}, that were previously used to build anomaly detection AIOps solutions, namely Yahoo S5\footnote{Yahoo Data Source: \url{https://yahooresearch.tumblr.com/post/114590420346/a-benchmark-dataset-for-time-series-anomaly}} and the Numenta Anomaly Benchmark (NAB)\footnote {NAB Data Source: \url{https://github.com/numenta/NAB/tree/master/data}}. Although the publicly available benchmarks received criticism in the way they are labeled~\cite{eamonn}, labeling anomalies in real-world AIOps datasets is complex, labor-intensive, and prone to error due to manual labeling and the subjectivity of the annotator~\cite{aiopsarxiv}. For instance, some annotators correlate anomalies with the moment an incident occurs~\cite{aiopsarxiv}, without taking into account that the points before and after the incident are identical as also pointed out by~\cite{eamonn}. Furthermore, currently, the AIOps time series data are collected from various sources, which is the reason why anomalies exhibit different behaviors and they are not consistent across time-series~\cite{aiopsarxiv}. Therefore, the anomaly detection AIOps datasets benchmarks reflect real-world characteristics of operational data, but as mentioned in~\cite{aiopsarxiv},~\cite{eamonn}, the labeling process requires more transparency by understanding the human decision-making process while labeling and the cause of the anomaly.

\textbf{Yahoo} is a dataset released by Yahoo Lab which contains both synthetic and real data collected from \textit{internet traffic} generated by Yahoo services. In our experiments, we solely considered real-world data (the Yahoo A1 benchmark), which we are further referring to as Yahoo. This dataset is composed of 67 time series collected with a granularity of 1 hour for 31 days (the shortest time series) and 61 days (the longest time series). The anomalous and non-anomalous points are labeled by domain experts.


\textbf{NAB} is a publicly available dataset previously used as a benchmark to assess the performance of anomaly detection models~\cite{nabbenchmark}. The NAB corpus contains multiple types of time series, such as metro traffic, tweets, etc. For our experiments, we solely select datasets referring to operational data, namely the realAWSCloudwatch time series that contain real-world server metrics (e.g. CPU utilization, Network Bytes In, Disk Read Bytes) collected by the AmazonCloudwatch service. For simplicity, we are further referring to these time series as NAB. This dataset is composed of 17 time series collected with a granularity of 5 minutes for 5 days (the shortest time series) and 17 days (the longest time series). To obtain the labels, we employ the ground truth labeling approach (labeling only the ground truth as an anomaly) instead of labeling the entire region to avoid raising too many false alarms to engineers~\cite{Ban2021CombatSA}.


\begin{table}[ht]
\begin{center}
  \caption{\label{table_datasets} Overview of the datasets characteristics.}
\begin{tabular}{|c|c|c|c|}
\hline
\multicolumn{1}{|c|}{Dataset} & \multicolumn{1}{c|}{NAB} & \multicolumn{1}{c|}{Yahoo A1} \\
\cline{2-3}
\hline
Number of Time Series in Dataset   & 17     & 67 \\
Granularity   & 5 min     & 1 hour \\
Approx. Min Data Collection Time & 5 days &   31 days \\
Approx. Max Data Collection Time & 17 days &  61 days \\
\hline
\end{tabular}

 \end{center}
\end{table}

\paragraph{Data-Splitting And Preprocessing}
In our study, we considered the same data-splitting scenario from previous works \cite{anomalyDetectionMicrosoft}, \cite{ajointmodelforit}, namely for each time series from each dataset the first half is used for training and the second half is used for testing. For our experiments, we assume that the anomaly labels are known for the training set.

\subsection{Models}
\subsubsection{State-of-the-Art Anomaly Detection}
We replicate five popular unsupervised anomaly detection models based on signal reconstruction, namely \textit{FFT}~\cite{fft}, \textit{SR}~\cite{anomalyDetectionMicrosoft}, \textit{PCI}~\cite{pci}, \textit{LSTM-AE}~\cite{lstmae}, and \textit{SR-CNN}~\cite{anomalyDetectionMicrosoft}. We selected our models based on both their popularity and their dissimilarity in techniques to detect anomalies. Furthermore, these models were previously used for anomaly detection in AIOps \cite{anomalyDetectionMicrosoft}, \cite{ajointmodelforit}, \cite{anomalydetectionAlibaba}. Therefore, we employ three models that work by solely applying different mathematical operations of the given time series to detect anomalies (FFT, PCI, and SR) and two models that require learning the behavior of the time series (LSTM-AE, and SR-CNN). 

The \textbf{FFT} method firstly transforms the time series from the time domain into the frequency domain using the Fourier transform and then transforms the time series back into the time domain to estimate a fitted curve of the data. Anomalies are detected by finding differences between data points and the fitted curves. In our study, we use the implementation of FFT from the original paper~\cite{fft}, which is publicly available\footnote{FFT Open Source Implementation: \url{https://github.com/HPI-Information-Systems/TimeEval-algorithms/tree/main/fft}}. Given the unavailability of a threshold that differentiates anomalies from non-anomalies, we empirically computed the threshold using the training data. Therefore, we remove all existing anomalies from the training data and we extract the maximum anomaly score on the training data. Sample with a higher anomaly score than the threshold are classified as anomalies.

The \textbf{SR} method is composed of three steps: the Fast Fourier Transform, which transforms the time series from the time domain into the frequency domain to calculate the log amplitude spectrum, the spectral residuals calculation and the Inverse Fourier Transform transforms the signal from the frequency domain back into the time domain. After these three steps, the saliency map is computed and used to detect anomalies through a threshold. In our study, we use the publicly available implementation of SR\footnote{SR Open Source Implementation: \url{https://github.com/y-bar/ml-based-anomaly-detection}} from the original paper~\cite{anomalyDetectionMicrosoft}. Optimal hyperparameters are determined by performing a grid search starting from the parameters suggested in~\cite{anomalyDetectionMicrosoft} until similar results are obtained on Yahoo. The NAB dataset is not assessed in the original paper and, thus, we perform a grid search to identify the optimal parameters.

The \textbf{PCI} method relies on the k-nearest neighbors of one data point to identify if that specific data point is an anomaly. The k-nearest neighbors are the closest continuous data points to the targeted data point in the time series. After calculating the nearest neighbors, the prediction confidence interval is calculated. The current data point is classified as an anomaly if its value is outside of the prediction confidence interval. We use the publicly available implementation of PCI\footnote{PCI Open Source Implementation: \url{https://github.com/HPI-Information-Systems/TimeEval-algorithms/tree/main/pci}} from the original paper~\cite{pci}. Optimal parameters are computed for each dataset through grid search.

The \textbf{LSTM-AE} method uses an autoencoder architecture to learn the representation of a time series in the time domain. An anomaly is detected when the error between the reconstructed and true time series is higher than a predefined threshold. Due to the lack of publicly available implementations of LSTM-AE, we implement it based on the architecture presented in~\cite{ubertimeseries}. We remove anomalous samples from the training set~\cite{ajointmodelforit} and perform interpolation between the closest non-anomalous samples is performed to preserve the continuity of the time series. Our implementation is publicly available in our replication package\footnote{Replication Package: \url{https://github.com/LorenaPoenaru/anomaly\_detection}}.

The \textbf{SR-CNN} method is composed of two main parts: the synthetic data generation and the CNN. It uses the saliency map calculated through spectral residuals as features for the CNN. The synthetic data generator injects fake anomalous points into the saliency map to  reduce the high imbalance between anomalous and non-anomalous points. The augmented data is used to train the CNN and distinguish between anomalies and non-anomalies. We used the publicly available implementation\footnote{SR-CNN Open Source Implementation: \url{https://github.com/microsoft/anomalydetector}} of SR-CNN from the original paper~\cite{anomalyDetectionMicrosoft}. We perform a grid search to determine the optimal parameters of SR-CNN on both datasets.

\subsubsection{Testing Window Size Assessment} Previous work reported results by splitting the original time series into half and using the entire first half for training and the entire second half for testing. We are further referring to this experimental setup as the SoTA setup. However, since we aim to understand the effect of retraining techniques on the available datasets, we also need to ensure that the model's predictions are not impacted by a smaller testing window. Thus, we split the testing sets into smaller subsets, and we evaluate the same model on each of the subsets. We further combine the predictions on the subsets and compare them to the prediction obtained from the SoTA setup. We are further referring to this setup as the Window Size Setup. The testing subsets correspond to around 168 samples (one week) for Yahoo and 225 samples (one day) for NAB and they also represent the retraining periods. We use natural time intervals, one week and one day for Yahoo and NAB, respectively, since they are commonly used in model retraining for AIOps solutions~\cite{datasplittingdecisions},~\cite{nodefailureretrain},~\cite{nodefailureretrain2},~\cite{naturalintervalsaiops}. Moreover, these exact periods were chosen based on the amount of available data points in each time series (see Table \ref{table_datasets}). The same subsets are used in the model retraining experiments.

\subsection{Retraining Techniques}
\subsubsection{Retraining Data}
From a training data perspective, we compare three scenarios, namely the scenario when the model is never updated, which we are referring to as the \textit{static approach} and two scenarios in which the model is updated over time, namely the \textit{full history approach} and \textit{sliding window approach}. 

\begin{figure}
    \centering
    \includegraphics[width=0.5\textwidth]{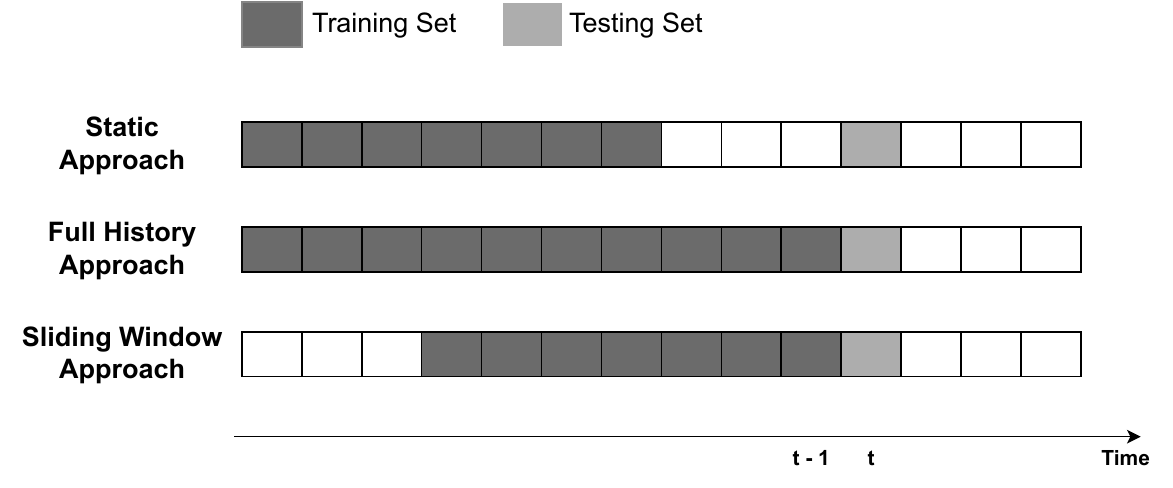}
    \caption{Training and testing data in case of the static approach, full-history approach, and sliding window approach.}
    \label{figure:retraining_approaches}
    \vspace{-1em}
\end{figure}

The testing data is split into equal testing batches corresponding to the predefined period. The experimental setup for the 3 scenarios can be observed in Figure \ref{figure:retraining_approaches}. According to each scenario, the model update is done as follows:

\textbf{Static Approach (S):} The model is trained only once on the training data and tested on the testing batches.

\textbf{Full History Approach (FH):} The model is retrained periodically by constantly enriching the training set. Thus, when testing the model on the testing set corresponding to batch \textit{t}, the model is trained on all the available data until batch \textit{t-1}.

\textbf{Sliding Window Approach (SW):} The model is retrained periodically by constantly replacing old data with new data. Thus, when testing the model on the testing set corresponding to batch \textit{t}, the model is retrained on only the newest data. In this approach, the training set size remains constant over time.

\subsubsection{Retraining Frequency}

\begin{figure}
    \centering
    \includegraphics[width=0.45\textwidth]{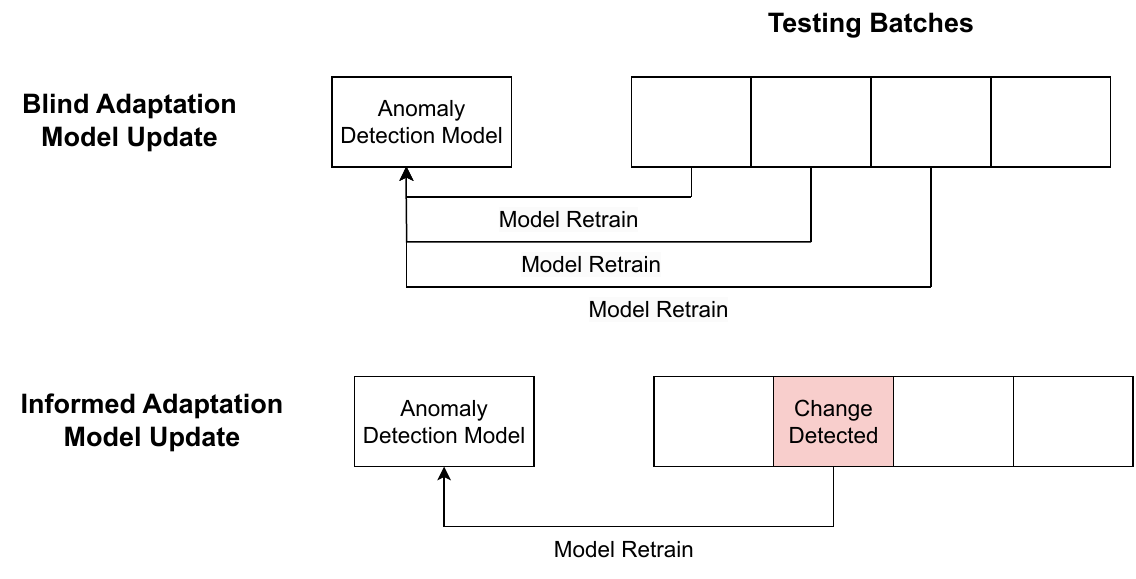}
    \caption{Blind vs Informed Retraining.}
    \label{figure:blind_informed}
    \vspace{-1em}
\end{figure}

From a retraining frequency perspective, we compare \textit{blind retraining} with \textit{informed retraining} updated models. The blind retraining technique implies that the model is retrained on a periodic basis by constantly including testing batches in the training data. Informed retraining implies that the model is retrained on all the data including the batch when the change occurs, only when a change detector identifies changes in data. A visual representation of the two is given in Figure \ref{figure:blind_informed}.

\subsection{Concept Drift Detectors}

For this study, we only consider FEDD as a concept drift detector since it was evaluated on both synthetic and real-world data~\cite{ipsoelmfedd}. For our experiments, we extract the implementation of FEDD from the replication package\footnote{ FEDD Open Source Implementation: \url{https://github.com/GustavoHFMO/IDPSO-ELM-S/blob/master/detectores/FEDD.py}} provided by the authors of~\cite{ipsoelmfedd}. Furthermore, we translate the code into English and include the FEDD implementation in our replication package together with an example of how to run it on any time series such that practitioners can use it for their data.

\subsection{Model Evaluation}
\subsubsection{Evaluation Metrics}
We use F1-score, Precision, and Recall to evaluate the performance of anomaly detectors. We do not consider the computation time as one of our evaluation metrics since the evaluated time series are relatively short and, thus, meaningful conclusions cannot be drawn. The Precision metric shows the ability of the anomaly detector not to label a non-anomaly as an anomaly and, therefore, not to raise too many false alarms. The Recall metric shows the ability of the anomaly detector to find all the anomalies and, therefore, to correctly identify all the anomalies. The F1-Score shows the compromise between Precision and Recall and, therefore, how many anomalies are correctly identified with respect to how many false alarms are triggered. The metrics are defined by the following equations:

\begin{equation}
  F1=\frac{2*Precision*Recall}{Precision+Recall}
  \label{eq1}
\end{equation}

\begin{equation}
  Precision = \frac{TP}{TP+FP}
  \label{eq2}
\end{equation}

\begin{equation}
Recall = \frac{TP}{TP+FN}
  \label{eq3}
\end{equation}

\noindent where TP and FN represent the number of True Positives and False Negatives, respectively. 

\subsubsection{Delay Metric} \label{evaluation_strategy}
\begin{figure}
    \centering
    \includegraphics[width=0.4\textwidth]{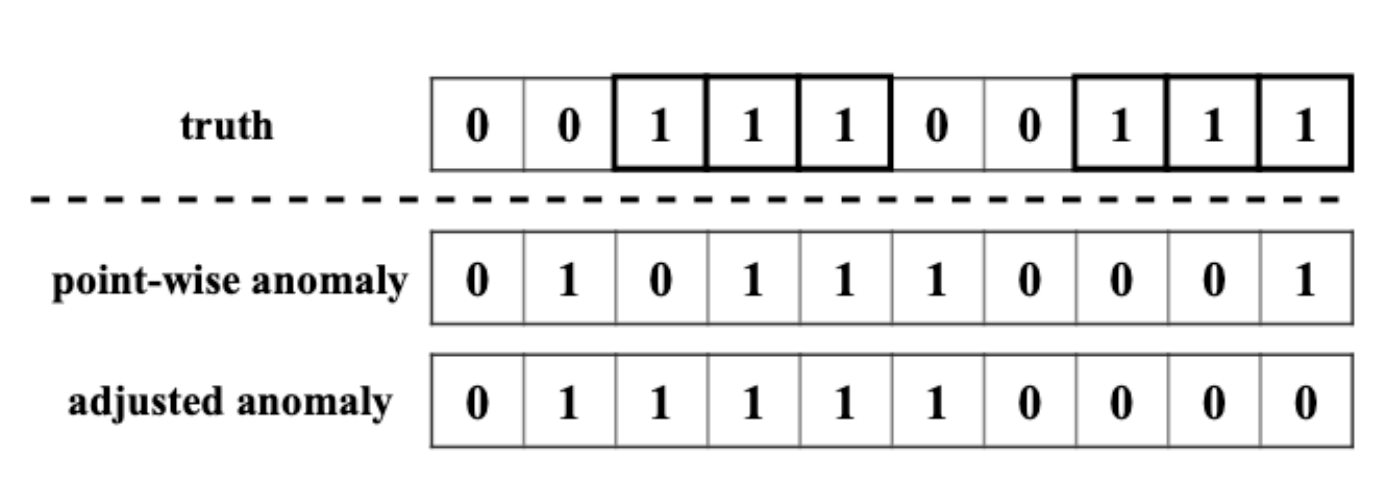}
    \caption{In this example we show the labels for 10 data points corresponding to one time series, where 1 indicates an anomaly and 0 indicates a non-anomalous point. The first row shows the ground truth, the second row shows the original predictions of one anomaly detection model and the third row shows the adjusted predictions considering a delay of 1. In the ground truth, there are 2 anomalous segments, each containing 3 anomalies. Since the model managed to predict the second anomaly in the sequence of anomalies and we tolerate a delay of 1 sample, the adjusted anomaly treats the entire first anomaly segment as a correct prediction. However, the second sequence of anomalies is treated as an incorrect prediction since even with a delay of one the anomaly on position 8 is not reported in time, while with a delay of 2, it would be reported in time. \cite{anomalyDetectionMicrosoft}}
    \label{figure:evaluation_strategy}
    \vspace{-1em}
\end{figure}

As aforementioned, AIOps anomaly detection models were evaluated considering a predefined prediction tolerance called delay. The detection delay is the maximum tolerable number of anomalous points in the anomaly segment that the detector can omit. If the detectors find any anomaly in the segment within this delay, the detection is considered correct. Otherwise, the detector fails to capture the anomaly. We depict in Figure~\ref{figure:evaluation_strategy} an example of metric computation including the delay.

\subsubsection{Statistical Test} We report the metrics averaged over all time series belonging to one dataset as done in previous work~\cite{anomalyDetectionMicrosoft},~\cite{unsupervisedADviaVAE},~\cite{ajointmodelforit}. To understand whether the differences between the averaged metrics are statistically significant we employ the Wilcoxon signed-rank statistical test and assess whether the difference between the metrics obtained for each time series in one scenario and the metrics obtained for each time series in another scenario is statistically significant. The Wilcoxon signed rank is a non-parametric statistical test, thus it does not make any assumption about the data distributions. Its null hypothesis is that the two populations analyzed come from the same distribution and, thus, there is no statistically significant difference between the two scenarios. We use a confidence interval of 90\% to assess whether the null hypothesis is accepted or rejected. To avoid bias in our experiments, we employ 5 random seeds in the case of LSTM-AE and SR-CNN, which suffer from randomness. We apply the statistical test on each random seed to assess the significance.

\section{Experimental Results}

\subsection{Anomaly Detection Models on Operational Data}
\subsubsection{State-of-the-Art Setup}
The first set of experiments aims to assess the performance of state-of-the-art anomaly detection models on operational data. We initially train the model on the first half of each time series and test it on the second half (SoTA setup), similar to the state-of-the-art~\cite{anomalyDetectionMicrosoft},~\cite{unsupervisedADviaVAE},~\cite{ajointmodelforit}. 

\begin{table}[ht]
\centering
\caption{Results of anomaly detectors in SoTA setting. With \textbf{bold} we highlight the best scores. The obtained results are obtained by averaging the metric over all time series from each dataset and over the 5 random seeds.}
\begin{tabular}{| c | c | c | c | c | p{0.5cm} | p{0.5cm} | p{0.5cm} | p{0.5cm} | p{0.5cm} |} 

 \cline{2-5}
 \multicolumn{1}{c}{\textbf{}} &
 \multicolumn{1}{|c|}{\textbf{Model}} &
 \multicolumn{1}{|c|}{\textbf{F1-Score}}& 
 \multicolumn{1}{|c|}{\textbf{Precision}}& 
 \multicolumn{1}{|c|}{\textbf{Recall}}\\
 \cline{1-5}
 
\multirow{5}{0.2em}{\rotatebox[origin=c]{90}{\textbf{Yahoo}}} & FFT &  0.07 & 0.04 & 0.39\\
 & SR & \textbf{0.25} & \textbf{0.28} & \textbf{0.36} \\
 & PCI & 0.09 & 0.06 & 0.39 \\
&  LSTM-AE & \textbf{0.37} & \textbf{0.38} &  \textbf{0.49} \\
 & SR-CNN & \textbf{0.57} & \textbf{0.54} & \textbf{0.61} \\
 
  \cline{1-5}

 \multirow{5}{0.2em}{\rotatebox[origin=c]{90}{\textbf{NAB}}} & FFT & 0.04 & 0.02 & 0.12 \\
 & SR & 0.05 & 0.15 & 0.33  \\
 & PCI & 0.03 & 0.02 & 0.32 \\

& LSTM-AE & \textbf{0.31} & \textbf{0.37} & \textbf{0.44} \\
 & SR-CNN & \textbf{0.82} & \textbf{0.70} & \textbf{0.99} \\
\cline{1-5}
\end{tabular}
\label{table:experiment1}
\end{table}

We depict our findings in Table~\ref{table:experiment1} where we can observe that the performance of anomaly detection models is overall relatively low, which could be a reason for the labeling problems identified in~\cite{eamonn}. FFT and PCI models obtained the lowest F1-score for both datasets. When it comes to PCI, we can observe that it tends to classify most data points as anomalies given its high recall and low precision on both datasets. The same tendency can be observed when it comes to FFT on the Yahoo dataset and the SR on the NAB dataset. Given the low performance of FFT and PCI on both operational datasets, these two models are no longer considered in the following experiments. Moreover, the SR model is no longer considered in the following experiments on the NAB dataset for the same reason. The best-performing models on both datasets are LSTM-AE and SR-CNN.


\begin{figure*}[t]
\centering
\includegraphics[width=1\textwidth, ,height=0.28\textheight]{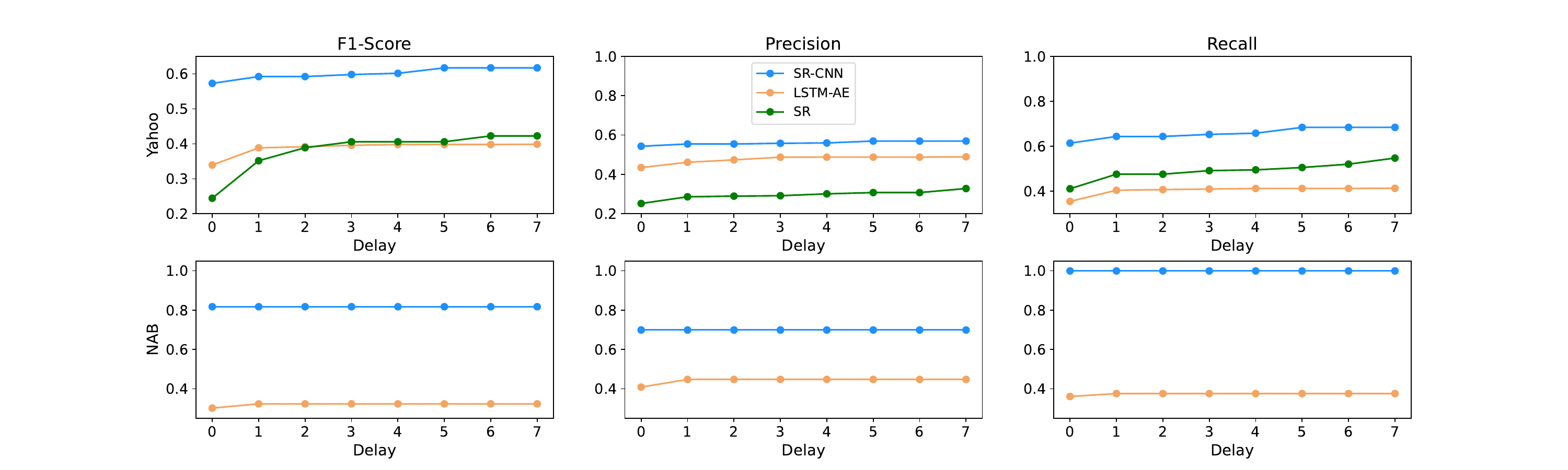}
\caption{Delay metric applied to F1-score, precision and recall on Yahoo and NAB.}
\label{figure:delay_new}
\end{figure*}

In Table~\ref{table:experiment1} we show the results of each model when performing a strict evaluation, where no delay is tolerated (delay=0). Thus, the model is penalized when it does not manage to detect the anomaly at the exact moment when it occurs. As aforementioned, AIOps models were evaluated in previous studies~\cite{anomalydetectionAlibaba},~\cite{anomalyDetectionMicrosoft},~\cite{ajointmodelforit} with a maximum delay of 7 points. Therefore, in Figure~\ref{figure:delay_new} we show the results for the same models with a delay of up to 7 points.

From Figure~\ref{figure:delay_new} we can notice that although on the Yahoo dataset, a higher delay can significantly influence the performance of anomaly detection models, on the NAB dataset it has almost no impact. In the case of Yahoo, the explanation behind this result is that 73.13\% of the time series belonging to the Yahoo dataset contain sequences of anomalies (group anomalies). When it comes to NAB, only 29.41\% of the time series contain sequences of anomalies. Therefore, the dataset containing samples related to internet traffic (Yahoo) contains significantly more group anomalies compared to the dataset containing samples related to CPU utilization (NAB), which shows the different behavior of different operational datasets.

Another important observation is that in the case of Yahoo, the SR model performs better after a delay of 2 points compared to LSTM-AE. However, if we assess the two anomaly detection models according to how well they detect the exact moment anomalies occur, the difference in the F1-score is almost 10\%. In our work, we aim to understand how well these methods perform when detecting the exact anomalous point is crucial. Therefore, for the rest of our experiments, we are solely considering a delay of 0 points when evaluating different anomaly detection models.

\begin{table}[ht]
\centering
\caption{Assessment of anomaly detectors' robustness towards testing window size (Window Size Setup). The obtained results are obtained by averaging the metric over all time series included in one dataset and over the 5 random seeds.}
\begin{tabular}{| c | c | c | c | c | p{0.5cm} | p{0.5cm} | p{0.5cm} | p{0.5cm} | p{0.5cm} |} 

 \cline{2-5}
 \multicolumn{1}{c}{\textbf{}} &
 \multicolumn{1}{|c|}{\textbf{Model}} &
 \multicolumn{1}{|c|}{\textbf{F1-Score}}& 
 \multicolumn{1}{|c|}{\textbf{Precision}}& 
 \multicolumn{1}{|c|}{\textbf{Recall}}\\
 \cline{1-5}
 
\textbf{Yahoo}
& LSTM-AE & 0.37 & 0.38 & 0.49 \\
& SR-CNN & 0.14 & 0.14 & 0.14 \\
 
  \cline{1-5}

\textbf{NAB}

& LSTM-AE & 0.31 & 0.37 & 0.44 \\
 & SR-CNN & 0.05 & 0.05 & 0.05\\
\cline{1-5}
\end{tabular}
\label{table:experiment1_window}
\end{table}

\subsubsection{Window Size Setup}
To fully answer our first research question, we need to understand whether models are robust towards the testing data size. We, therefore, employ the Window Size Setup described in Section 5.2.2 in the case of SR-CNN and LSTM-AE and compare it to the performance of the model when using the SoTA setting. However, as part of the functionality of SR, this model requires to be evaluated on a continuous time series. Therefore, this model could only be assessed for robustness on the sub-testing set immediately following the training set. From this experiment, we noticed that SR is not influenced by the size of the testing lengths and, thus, is robust to changes in the testing length. For SR-CNN and LSTM-AE we show results in Table~\ref{table:experiment1_window}.


Throughout our experiments, we noticed that the performance of LSTM-AE in both scenarios is the same. Therefore, the testing window size does not influence this anomaly detector. As it can be observed from Table~\ref{table:experiment1_window}, SR-CNN is drastically affected by being evaluated on smaller window sizes on both datasets. Its performance drops from an F1-score of 0.82 to 0.05 on the NAB dataset and from an F1-score of 0.57 to 0.14 for the Yahoo dataset.Thus, we consider that only SR and LSTM-AE are robust toward different testing sizes.

\subsection{Full-History vs Sliding Window}
In this set of experiments, we aim to answer our second research question and understand the differences in performance between the model that was never updated (S) compared to the models that were updated using a full-history (FH) vs a sliding window (SW) approach. In Table~\ref{table:experiment2} we show the results of this experiment. The results displayed for SR-CNN are the results obtained through the testing window size setup experiment since this experiment implies a division of the entire testing data into smaller subsets and retraining the model accordingly.

\begin{table*}[ht]
\centering
\caption{Results of anomaly detectors for blind model retraining. With \textbf{bold} we highlight the situation in which the difference in performance between the S model and the updated model (FH, SW) is significant. The displayed results are obtained by averaging the metric over all time series included in one dataset and over the 5 random seeds. (S-Static, FH-Full-History, SW-Sliding Window)}
\begin{tabular}{| c | c | p{0.5cm} | p{0.5cm} | p{0.5cm} | p{0.5cm} | p{0.5cm} | p{0.5cm} | p{0.5cm} | p{0.5cm} | p{0.5cm} |} 

 \cline{3-11}
   \multicolumn{2}{c}{\textbf{}} &
 \multicolumn{3}{|c|}{\textbf{F1-Score}}& 
 \multicolumn{3}{|c|}{\textbf{Precision}}& 
 \multicolumn{3}{|c|}{\textbf{Recall}}\\
 \cline{2-11}
   \multicolumn{1}{c}{\textbf{}} &
 \multicolumn{1}{|c}{\textbf{Model}} &
 \multicolumn{1}{|c}{\textbf{S}} &
 \multicolumn{1}{|c}{\textbf{FH}} &
 \multicolumn{1}{|c|}{\textbf{SW}} &
 \multicolumn{1}{|c}{\textbf{S}} &
 \multicolumn{1}{|c}{\textbf{FH}} &
 \multicolumn{1}{|c|}{\textbf{SW}} &
 \multicolumn{1}{|c}{\textbf{S}} &
 \multicolumn{1}{|c}{\textbf{FH}} &
 \multicolumn{1}{|c|}{\textbf{SW}}\\
 \cline{1-11}
 
\multirow{3}{0.1em}{\rotatebox[origin=c]{90}{\textbf{Yahoo}}} 

 & SR & 0.25 & 0.21 & \textbf{0.19} & 0.28 & \textbf{0.19} & \textbf{0.17} & 0.36 & \textbf{0.23} & \textbf{0.23}\\

& LSTM-AE & 0.37 & \textbf{0.42} & \textbf{0.46} & 0.38 & \textbf{0.42} & \textbf{0.46} & 0.49 & \textbf{0.53} & \textbf{0.59}\\
 & SR-CNN & 0.14 & 0.10  & 0.14 & 0.14 & 0.09 & 0.17 & 0.14 & \textbf{0.26} & 0.22\\
 
  \cline{1-11}

 \multirow{2}{0.1em}{\rotatebox[origin=c]{90}{\textbf{NAB}}} 
& LSTM-AE & 0.31 & \textbf{0.34} & \textbf{0.37} & 0.37 & 0.37 & \textbf{0.45} & 0.44 & \textbf{0.46} & \textbf{0.48}\\
 & SR-CNN & 0.05 & 0.07 & 0.03 & 0.05 &\textbf{0.15} & 0.07 & 0.05 & \textbf{0.13} & 0.04\\
\cline{1-11}
\end{tabular}
\label{table:experiment2}
\end{table*}

From Table~\ref{table:experiment2} we can notice that the performance of LSTM-AE is significantly improving when the model is updated over time. The highest performance is achieved when the model is updated using the sliding window (SW) technique. When it comes to the SR model, its performance degrades when being updated with the SW technique, but it does not statistically change when being updated with the FH technique. The SR-CNN performance benefits from being updated using the FH technique since it manages to find more anomalies (higher recall) and also to label more non-anomalies correctly (higher precision). In terms of SR, its performance degrades when old data is discarded. 

\subsection{Drift Detection based Retraining}

\subsubsection{Drift Detection Results}
\begin{figure}
    \centering
    \includegraphics[width=0.5\textwidth]{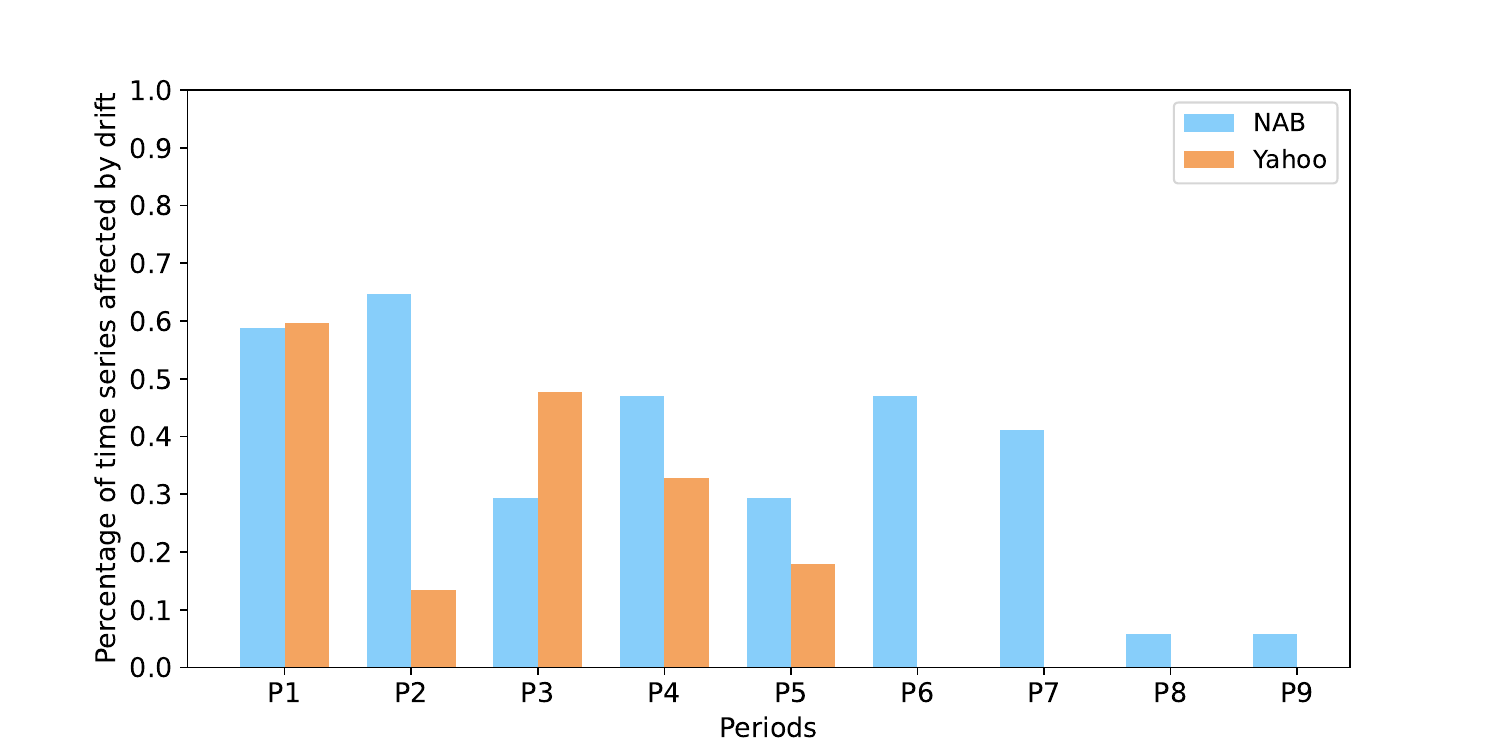}
    \caption{Percentage of time series affected by concept drift during each period according to FEDD.}
    \label{figure:drift_detection_results}
\end{figure}

We apply the concept drift detector FEDD on a period basis (weekly for Yahoo and daily for NAB) for each time series from each dataset. Figure~\ref{figure:drift_detection_results} shows the percentage of time series affected by concept drift according to FEDD during each period. Due to the lengths of time series corresponding to each dataset, the evaluation time series within the Yahoo dataset could be split into a maximum of five periods (five weeks), while those within the NAB dataset could be split into a maximum of nine periods (nine days).

From Figure~\ref{figure:drift_detection_results} we can see that both for Yahoo and NAB, more than 50\% of the time series are affected by concept drift in the first period (P1). Furthermore, in the case of NAB, during the second period (P2), there are more than 60\% of the time series affected by concept drift according to FEDD. 

\subsubsection{Blind vs Informed Retraining}
With this experiment, we aim to understand whether an anomaly detection model's performance can be preserved when retraining based on a drift detector's output (informed retraining) and how it compares to periodically retraining the model (blind retraining). Since the most significant benefits of periodic model retraining were observed in the case of LSTM-AE, this experiment is solely performed on this anomaly detection model. In addition, this experiment is performed with both data retraining techniques, full history (FH) and sliding window (SW).

\begin{table}[ht]
\centering
\caption{Results of LSTM-AE when comparing blind with informed model retraining. The displayed results are obtained by averaging the metric over all time series included in one dataset and over the 5 random seeds. \textbf{Bold} means that informed retraining achieves significantly better results compared with the static model. \textit{Italic} means that informed retraining achieves significantly similar results with blind retraining. (S-Static, FH-B/I-Full History Blind/Informed, SW-B/I-Sliding Window Blind/Informed.)}
\begin{tabular}{| c | c | c | c | c | c | c |}

 \cline{3-7}
   \multicolumn{1}{c}{\textbf{}} &
 \multicolumn{1}{c}{\textbf{ }} &
 \multicolumn{1}{|c}{\textbf{S}} &
 \multicolumn{1}{|c}{\textbf{FH-B}} &
 \multicolumn{1}{|c|}{\textbf{FH-I}} &
 \multicolumn{1}{|c}{\textbf{SW-B}} &
 \multicolumn{1}{|c|}{\textbf{SW-I}}\\
 \cline{1-7}
 
\multirow{3}{0.1em}{\rotatebox[origin=c]{90}{\textbf{Yahoo}}}

 & F1 & 0.37 & 0.42 & \textbf{0.40} & 0.46 & \textbf{0.41} \\
 & Precision & 0.38 & 0.42 & \textbf{0.40} & 0.46 & \textbf{\textit{0.45}}\\
 & Recall & 0.49 & 0.53 & 0.50 & 0.59 & \textbf{0.55}\\

 \cline{1-7}

 \multirow{3}{0.1em}{\rotatebox[origin=c]{90}{\textbf{NAB}}} 
 
 & F1 & 0.31 & 0.34 & \textbf{\textit{0.34}} & 0.37 & \textbf{\textit{0.36}} \\
 & Precision & 0.37 & 0.37 & \textbf{0.39} & 0.45 & \textbf{\textit{0.45}}\\
 & Recall & 0.44 & 0.46 & 0.45 & 0.48 & \textbf{0.46}\\
 
 \cline{1-7}
\end{tabular}
\label{table:experiment3}
\end{table}

Table~\ref{table:experiment3} shows that overall the performance of informed model retraining is higher than the performance of a model that was never updated (static model). However, blind (periodic) model retraining generally achieves better results than informed model retraining. Similar to the periodic model retraining experiment, the highest results were achieved using the sliding window approach.

From Table~\ref{table:experiment3} it can be noticed that the informed model retraining achieved similar precision to the blind model retraining for both datasets when the model was retrained using the sliding window approach. However, in all situations informed retraining results in lower recall compared with blind model retraining. When it comes to the full-history approach retraining technique, all performance metrics calculated for the blind model retraining are always higher than the performance metrics calculated for the informed model retraining on the Yahoo dataset. However, for the NAB dataset, the F1 score for informed model retraining is similar to the blind model retraining, while the precision is slightly higher.

\section{Discussions and Answers to Research Questions}

\fbox{\begin{minipage}{23em}
\par{\textbf{\textit{Research Question 1:}} What is the performance of state-of-the-art anomaly detection models on operational data and how does the testing window size influence it?}
\par{\textbf{\textit{Answer 1:}} The more complex models (LSTM-AE and SR-CNN) perform significantly better on both operational datasets compared to the simpler models (FFT, PCI, SR). However, the evaluation window size significantly impacts SR-CNN but does not affect LSTM-AE or SR.}
\end{minipage}}

\phantom{r}
\textbf{State-of-the-Art Anomaly Detection:} Throughout our experiments, we noticed that PCI and FFT tended to label every point in the time series as an anomaly. This  shows the complexity of operational data also described in~\cite{aiopsarxiv}, which was only captured by LSTM-AE and SR-CNN. Moreover, we noticed a significant difference of almost 10\% between LSTM-AE and SR anomaly detectors when the delay metric is considered. We believe that the delay could be misleading for practitioners who want to use the anomaly detection models and, thus we recommend researchers to initially report the results with a delay of 0 and afterward perform an additional experiment with a variable delay.

One interesting finding of our experiment is that SR-CNN is highly influenced by the size of the data it is evaluated on. When performing the experiments to assess this aspect, we noticed a significant decrease in performance when comparing the SR-CNN tested on the original setup described in the paper~\cite{anomalyDetectionMicrosoft} to a setting in which the testing set is smaller. However, in both settings, we respected the original SR-CNN architecture in which the training and testing datasets are separated before performing the time series spectral transformation. This could be due to the functionality of SR-CNN since this model relies on decomposing time series into another domain and extracting features that are further used to train a CNN. Therefore, when shortening the time series, these features might not be significant enough for CNN to properly distinguish between anomalies and non-anomalies. 

\textbf{Implications for Practitioners:} In the light of our findings we strongly recommend practitioners that when performing anomaly detection model selection they should consider a delay of 0. Thereafter, they can adjust the delay according to the requirements of their application. Given our findings regarding the performance drop of SR-CNN when a smaller testing size is employed, we further suggest that during model selection, AIOps practitioners should consider its robustness to different time series lengths and how often the inference is going to be performed. For instance, when identifying anomalies in a time series with a big granularity (e.g. 1 hour) that needs to be evaluated daily, employing an anomaly detector that is sensitive to short testing sets could lead to erroneous results. Moreover, sometimes incidents could occur and data collection could be interrupted, which can also shorten the available testing data.

\phantom{r}

\fbox{\begin{minipage}{23em}
\par{\textbf{\textit{Research Question 2:}} What is the impact of the two model retraining techniques from the perspective of the retraining data (full-history vs sliding window approach)?}
\par{\textbf{\textit{Answer 2:}} The sliding-window approach is beneficial to models that learn from the time domain (LSTM-AE) but can impact the performance of models decompose the time series into another domain (SR and SR-CNN). Retraining using the full-history approach achieved higher performance than retraining in most of the cases.}
\end{minipage}}

\phantom{r}

We empirically prove that the LSTM-AE anomaly detector benefits from being periodically updated in both datasets, being able to identify more anomalies and distinguish better between anomalous and non-anomalous points. The performance of this model is higher when employing the sliding window approach than the full-history approach. This could show that old samples are no longer relevant to the current behavior of the data and retraining the LSTM-AE only on the most recent data eliminates non-relevant data points. When it comes to SR, which is not affected by the window size as shown in previous experiments, we noticed that a sliding window approach significantly impacts its performance. This can be a consequence of lowering the size of the time series and discarding important spectral information. 

\textbf{Implications for Practitioners:} Based on our findings, periodically retraining anomaly detection models can significantly improve their performance over time. Thereby, AIOps practitioners should consider periodic model maintenance after deployment as part of the anomaly detection model lifecycle. Furthermore, we noticed that models that identify anomalies from the original time series (LSTM-AE) benefit from a sliding-window retraining approach, while the ones that detect anomalies by transforming the time series in other domains (SR) benefit from a full-history approach. Therefore, the retraining technique should be chosen according to the employed anomaly detector.

\phantom{r}

\fbox{\begin{minipage}{23em}
\par{\textbf{\textit{Research Question 3:}} What is the impact of the two model retraining techniques from the perspective of the retraining frequency (blind vs informed retraining)?}
\par{\textbf{\textit{Answer 3:}} Blind (periodic) retraining usually achieves higher performance than informed retraining. However, when a sliding window retraining technique is employed, the precision is statistically similar.}
\end{minipage}}

\phantom{r}

This experiment shows that retraining a model based on the output of a drift detector (FEDD) achieves a higher performance than a model that was never updated. This shows that the informed retraining setting is a promising research path and the FEDD drift detector could be employed as an anomaly detection model degradation monitoring tool. However, a model that is retrained periodically achieves slightly higher performance than a model retrained based on the output of a drift detector. This could be caused by the fact that not all drifts are captured by FEDD, which leaves room for improvement of concept drift detectors for time series. 

We noticed from our experiments that for both datasets the precision obtained through blind retraining is comparable to the one obtained through informed retraining when a sliding window approach is employed. This shows that retraining based on a drift detector helps reduce the number of false alarms triggered by an anomaly detector as much as periodic retraining does. However, with a periodic model retraining more anomalies are captured compared to the blind model retraining given the higher recall. 

\textbf{Implications for Practitioners:} Since we show the potential of having a model retrained based on the output of a drift detector, we argue that these monitoring tools can be employed in real-world anomaly detection model maintenance pipelines. Furthermore, to understand the suitability of the employed drift detector and the effects of retraining based on its output, we suggest that practitioners allocate a period to evaluate the model quality over time. Within this period AIOps practitioners should constantly investigate the effects of different retraining techniques (blind vs informed and sliding-window vs full-history) and understand which maintenance technique is the most suitable for anomaly detection AIOps solutions with respect to its corresponding costs (labeling/labor costs, etc). Within this period multiple drift detectors can be evaluated .


\section{Conclusions and Future Work}
In this paper, we investigate different maintenance techniques for popular anomaly detection models on two AIOps domains, internet traffic and CPU utilization. From a retraining frequency perspective, we analyzed \textit{blind retraining}, retraining the model periodically, and \textit{informed retraining}, retraining the model when a drift detector indicates it. With this study, we assess the potential of using drift detectors as a quality monitoring tool for anomaly detection models. From a retraining data perspective, we experimented with \textit{full-history approach}, constantly enriching the training set, and \textit{sliding window approach}, discarding old samples from the training set. 

We began our study by replicating the most popular anomaly detection model and testing their performances on the two AIOps datasets. We observed that some models could not detect anomalies in this type of data, while others were sensitive to the length of the testing data size. Our study shows that generally, an updated anomaly detector achieves higher performance than an anomaly detector that was never updated. We observed that the sliding-window approach benefits models that identify anomalies from the original time series while the full-history approach benefits models that transform time series into another domain to detect anomalies. Furthermore, we empirically demonstrated that the performance of an anomaly detector retrained based on the output of a drift detector leads to better performance when compared to an anomaly detector that was never updated. This shows that drift detection-based model update is a promising research path and can be the beginning of automated model maintenance pipelines. In light of our findings, we offer recommendations to AIOps practitioners regarding anomaly detection model selection and assessment, as well as identifying the most appropriate model maintenance techniques.

Our work was limited by the availability of open-source AIOps datasets for time series anomaly detection. We, therefore strongly encourage AIOps practitioners to release more datasets that researchers can use as benchmarks when developing new models. Moreover, we also believe that research should focus on understanding and reporting the labeling process of experts. This can offer anomaly detection researchers more transparency and confidence in the data quality.

In our study, we solely targeted anomaly detection for CPU utilization and internet traffic data. A promising research path is analyzing the potential of a concept drift detection-based framework in other AIOps applications, such as job and disk failure prediction~\cite{datasplittingdecisions}. Furthermore, we noticed that the performance of retraining an anomaly detector based on FEDD is lower than periodically updating the anomaly detector. This could show that FEDD does not capture all drifts that lead to performance degradation. Therefore, future work should consider investigating more drift detectors suitable for time series beyond FEDD and designing more accurate drift detectors for time series. 

\section{Threats To Validity}

\subsection{External Validity}

An external threat to validity derived from our study is the generalizability of our results towards different AIOps solutions. In our study, we targeted detecting anomalies in solely two AIOps domains, namely, internet traffic and CPU utilization due to public availability. Moreover, in this study, we relied on the provided anomalous labels corresponding to each dataset. The anomaly detection models depend on the datasets they are applied to. Although our datasets contain solely real-world data, the data used to e.g. predict incidents using AIOps solutions might be different than the one we employed. Therefore, the anomaly detection AIOps solutions lifecycle should be explored for other AIOps domains.

In our study, we included a variety of popular anomaly detection models, which are different from each other in terms of the technique they use to detect anomalies. Since we noticed that maintenance techniques are model-dependent when employing other models they need to be verified accordingly.

\subsection{Internal Validity}
When computing the static model, we used the first half of each time series from each dataset as training and the second half of each time series as testing as previous work \cite{anomalyDetectionMicrosoft}, \cite{ajointmodelforit}.


We partitioned our testing data into periods of different sizes (weekly and daily) similar to prior work \cite{datasplittingdecisions}, \cite{towardsconsistentinterpretationofAIOps}. We did not consider periods smaller than one day since not enough data is captured in a time shorter than one day. Periods bigger than one week were not considered given the size of the time series.


\subsection{Construct Validity}
We replicated anomaly detection models used in previous studies \cite{pci}, \cite{anomalyDetectionMicrosoft}, \cite{fft}. We employed hyperparameters used by previous studies. In the case when the hyperparameters were not provided, we used hyperparameters that led to similar results to what was reported in previous studies. In the case of LSTM-AE, we used the maximum reconstruction error on the training set as our anomaly threshold, which is derived from each time series. We employed multiple performance metrics to evaluate and interpret the model's outcome.




\printbibliography


\end{document}